\documentclass[lettersize,journal]{IEEEtran}
\usepackage{amsmath,amsfonts}
\usepackage{algorithmic}
\usepackage{algorithm}
\usepackage{array}
\usepackage[caption=false,font=normalsize,labelfont=sf,textfont=sf]{subfig}
\usepackage{textcomp}
\usepackage{stfloats}
\usepackage{url}
\usepackage{verbatim}
\usepackage{graphicx}
 \usepackage{multirow} 
 \usepackage{booktabs}
\usepackage{cite}
\usepackage{hyperref}
\usepackage{comment}
\usepackage{pifont}

\usepackage{arydshln}  
\begin{document}

\title{Sentinel2Cap: A Human-Annotated Benchmark Dataset for Multimodal Remote Sensing Image Captioning}

\author{Lucrezia Tosato$^{1,2}$,
        Gianluca Lombardi$^{3}$,
        and Ronny Hänsch$^{4}$%
\thanks{$^{1}$LIPADE, Université Paris Cité, 75006 Paris, France.}
\thanks{$^{2}$SAPIA, ONERA, Palaiseau, France.}
\thanks{$^{3}$LCQB, Sorbonne Université, CNRS, IBPS, 75005 Paris, France.}
\thanks{$^{4}$German Aerospace Center (DLR), We\ss ling, Germany (e-mail: ronny.haensch@dlr.de).}
\thanks{Lucrezia Tosato is now employed at SARMAP and can be reached at ltosato@sarmap.ch .}
\thanks{Manuscript received ..., 2026; revised ..., 2026.}}

\markboth{Journal of \LaTeX\ Class Files,~Vol.~14, No.~8, May~2026}%
{Shell \MakeLowercase{\textit{et al.}}: A Sample Article Using IEEEtran.cls for IEEE Journals}


\maketitle

\begin{abstract}




Image captioning has become an important task in computer vision, enabling models to generate natural language descriptions of visual content. While several datasets exist for natural images and high-resolution optical remote sensing imagery, the availability of captioning datasets for multimodal satellite data remains limited, particularly for SAR imagery and medium-resolution sensors. We introduce \textit{Sentinel2Cap}, a human-annotated multimodal captioning dataset containing Sentinel-1 SAR and Sentinel-2 multi-spectral image patches at 10 m and 20 m spatial resolution with diverse land cover compositions. Captions are created manually and carefully validated to ensure both semantic accuracy and linguistic quality. To evaluate \textit{Sentinel2Cap}, we perform a zero-shot captioning using the Qwen3-VL-8B-Instruct model across three image modalities: RGB, multi-spectral, and SAR pseudo-RGB representations. Results show that RGB images achieve the highest captioning performance, while SAR images remain more challenging for vision-language models. Providing modality-specific contextual prompts consistently improves performance across all metrics. These findings highlight both the challenges of multimodal remote sensing image captioning and the potential value of human-annotated datasets for advancing research in cross-modal scene understanding. All the material is publicly avaiable. 

\end{abstract}


\section{Introduction}





Image captioning aims to automatically generate natural language descriptions that summarize the content of an image. In recent years, advances in deep learning and the development of large-scale vision-language models (VLMs) have significantly improved caption generation for natural images. Benchmark datasets such as COCO Captions~\cite{chen2015microsoft} have played a central role in this progress by providing large collections of images paired with high-quality human-written descriptions.

In remote sensing, image captioning has gained increasing attention as a way to improve scene interpretation, support information retrieval, and facilitate interaction between geospatial data and natural language systems. However, most existing remote sensing captioning datasets focus on high-resolution optical imagery and are often limited in size or domain coverage. Furthermore, only a small number of datasets include Synthetic Aperture Radar (SAR) data, despite its importance for Earth observation due to its ability to operate independently of daylight and weather conditions.

Another limitation of current datasets is that many rely on automatically generated captions, which may introduce linguistic inconsistencies or semantic inaccuracies. Human-written captions remain valuable for capturing detailed spatial relationships, contextual information, and natural language diversity. At the same time, relatively little work has explored captioning datasets that combine multiple sensing modalities, such as optical and SAR imagery, particularly at medium spatial resolution.

\begin{figure*}[!t]
\centering
\includegraphics[width=\textwidth]{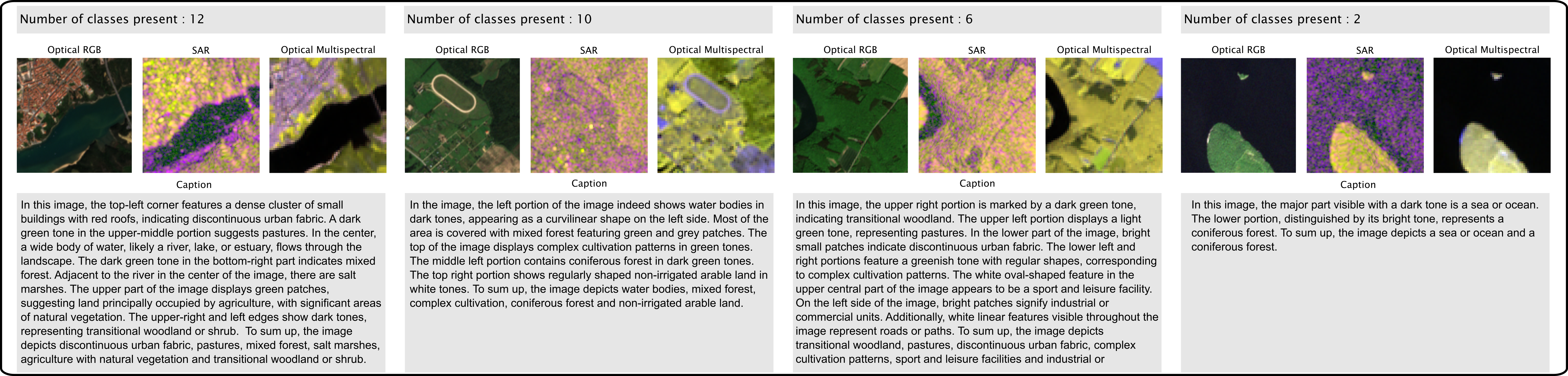}
\caption{Examples from \textit{Sentinel2Cap} showing multimodal image patches (MSI, SAR, and RGB) with decreasing semantic complexity (12, 10, 6, and 2 land-cover classes) and their corresponding human-annotated captions, illustrating the dataset’s emphasis on semantically rich scenes and high-quality manual annotation.}
\label{fig:dataset1}
\end{figure*}

To address these limitations, we introduce \textit{Sentinel2Cap}, a multimodal remote sensing image captioning dataset built on the Refined BigEarthNet (reBEN) dataset~\cite{clasen2024reben}. The dataset contains Sentinel-1 SAR and Sentinel-2 multi-spectral image patches covering multiple European regions and representing a wide range of land cover types. We encourage semantic richness by prioritizing scenes containing multiple classes during image selection. Captions are manually created by human annotators followed by a structured validation process to ensure both semantic correctness and grammatical quality.
Some examples of \textit{Sentinel2Cap} are presented in Figure~\ref{fig:dataset1}. 

We conduct a baseline evaluation using the Qwen3-VL-8B-Instruct vision-language model in a zero-shot setting. Captioning performance is evaluated across three image modalities: RGB, multi-spectral, and SAR pseudo-RGB representations. The evaluation investigates how modality and prompt design influence caption generation quality.

All the material is publicly avaiable at \url{https://github.com/LucreziaT/Sentinel2Cap}.

\section{Related Work}

\subsection{Captioning Datasets in Computer Vision}

Image captioning has long been a central task in the computer vision community, supported by the development of large-scale datasets that pair images with natural language descriptions. One of the most widely used benchmarks is COCO Captions~\cite{chen2015microsoft}, which contains more than 330,000 images annotated with approximately half a million human-written captions. Each image is associated with five diverse sentences, averaging around ten words, collected through crowd-sourcing to ensure both linguistic richness and descriptive relevance.

Several datasets have been proposed to address limitations of this benchmark and explore more challenging captioning scenarios. The nocaps dataset~\cite{agrawal2019nocaps}, for instance, extends the evaluation of captioning systems to an open-vocabulary setting. It contains 15,100 images paired with 166,100 human-written captions, with each image annotated by eleven sentences to encourage models to generalize to objects not observed during training.

Other datasets focus on more specialized domains. Fashion IQ~\cite{wu2021fashion} targets the fashion domain and provides human-written sentence pairs describing fine-grained attribute differences between visually similar items. Rather than focusing on general scene descriptions, the dataset facilitates text-based image retrieval by emphasizing subtle variations in appearance and style.

Automated caption generation has also been explored in video-based settings. The Soccer Captioning dataset~\cite{hammoudeh2022soccer}, for example, contains captions generated from more than 500 hours of SoccerNet videos, producing approximately 22,000 samples centered on action recognition and sports commentary. While this dataset enables large-scale annotation, the resulting captions are typically more formulaic and exhibit less linguistic variability compared to human-written descriptions.

Despite the progress enabled by these datasets, they are primarily designed for conventional RGB imagery captured from ground-level perspectives. As a result, they do not address the specific challenges posed by remote sensing data, such as overhead viewpoints, large variations in spatial scale, and the complex spatial organization of geospatial scenes.

\subsection{Captioning Datasets in Remote Sensing}

\begin{table*}[t]
\centering
\caption{Overview of remote sensing captioning datasets. \ding{51} / \ding{55} indicate the availability of SAR, optical data, and human-written captions. \textit{Sentinel2Cap} is highlighted as the only dataset combining multimodal SAR+optical data with human-annotated captions at medium resolution.}
\resizebox{\textwidth}{!}{
\begin{tabular}{l l c c c l l l}
\hline
\textbf{Name (Citation)} & \textbf{Task} & \textbf{SAR} & \textbf{Optical} & \textbf{Human} & \textbf{Resolution} & \textbf{\# Images} & \textbf{Captions / Img} \\
\hline

\multicolumn{8}{l}{\textit{RGB / Optical Captioning Datasets}} \\

NWPU-Captions~\cite{cheng2022nwpu} 
& Captioning & \ding{55} & \ding{51} & \ding{51} & Varied ($\sim$256 px) & 31,500 & 5 \\

UCM-Captions~\cite{yamani2024remote} 
& Captioning & \ding{55} & \ding{51} & \ding{51} & 0.3 m & 2,100 & 5 \\

Sydney-Captions~\cite{yamani2024remote} 
& Captioning & \ding{55} & \ding{51} & \ding{51} & Not specified & -- & 5 \\

RSICD~\cite{lu2017exploring} 
& Captioning & \ding{55} & \ding{51} & \ding{51} & Varied & 10,921 & 5 \\

RSITMD~\cite{yuan2022exploring} 
& Retrieval & \ding{55} & \ding{51} & \ding{51} & 0.3--1 m & 4,743 & -- \\

LEVIR-CC~\cite{liu2022remote} 
& Change Captioning & \ding{55} & \ding{51} & \ding{51} & 1024$\times$1024 & 10,077 pairs & 5 \\

XLRS-Bench~\cite{wang2025xlrs} 
& Captioning & \ding{55} & \ding{51} & \ding{55} & Ultra-high & 1,400 & $\sim$0.7 \\

\hline

\multicolumn{8}{l}{\textit{SAR Captioning Datasets}} \\

SAR Ship Captioning~\cite{zhao2022exploring} 
& Captioning & \ding{51} & \ding{55} & \ding{51} & Not specified & 1,500 & 2 \\

SARLANG-1M~\cite{wei2025sarlang} 
& Captioning & \ding{51} & \ding{51} & \ding{55} & Not specified & 1M & $\ll$1 \\

SAR-TEXT~\cite{cheng2025sar} 
& Captioning & \ding{51} & \ding{55} & \ding{55} & Multi-res & $>$130k & 1 \\

\hline

\multicolumn{8}{l}{\textit{Multimodal Captioning Dataset}} \\
BigEarthNet.txt~\cite{herzog2026bigearthnet} 
& Captioning / VQA
& \ding{51} 
& \ding{51} 
& \ding{55}
& Medium (10--20 m) 
& 464,044 
& $\sim$20+ \\
GAIA~\cite{zavras2026gaia} 
& Captioning
& \ding{51} 
& \ding{51} 
& \ding{55} 
& Multi-scale 
& 40,201 
& 5 \\
\textbf{\textit{Sentinel2Cap} (Ours)} 
& \textbf{Captioning} 
& \textbf{\ding{51}} 
& \textbf{\ding{51}} 
& \textbf{\ding{51}} 
& \textbf{Medium (10--20 m)} 
& \textbf{12000} 
& \textbf{1} \\

\hline
\end{tabular}
}
\label{tab:rs_caption_datasets}
\end{table*}

In recent years, remote sensing image captioning has attracted increasing attention, with the number of publications in this area growing by nearly 50\% between 2023 and 2024~\cite{zhang2024review}. To support this development, several datasets have been introduced, most of which focus on high-resolution RGB imagery captured from aerial or satellite platforms.

One of the most widely used benchmarks in this domain is the NWPU-Captions dataset~\cite{cheng2022nwpu}, derived from NWPU-RESISC45~\cite{cheng2017remote}. It contains 31,500 RGB optical images annotated with 157,500 human-written captions. The annotation process follows strict guidelines requiring the inclusion of all prominent objects, discouraging vague or directional expressions, and promoting syntactic diversity. Each caption must contain at least six words, and annotators are encouraged to avoid ambiguities while ensuring that the description captures the key visual content of the scene.

Other datasets provide similar annotations on smaller collections of aerial imagery. UCM-Captions~\cite{yamani2024remote} and Sydney-Captions~\cite{yamani2024remote} both include five captions per image that describe the same scene using different wording. UCM-Captions consists of 2,100 RGB images with a spatial resolution of 0.3 meters, while Sydney-Captions is built from Google Earth imagery covering the Sydney metropolitan area.

The RSICD dataset~\cite{lu2017exploring} further expands the available training data with 10,921 images and 54,605 captions collected from multiple sources, including Google Earth, Baidu Map, MapABC, and Tianditu. Each image is associated with five descriptions, although in some cases slight variations of the same sentence are used to meet the annotation requirements. Similar to other datasets, the annotation process discourages vague adjectives and repetitive sentence structures in order to maintain linguistic diversity while preserving semantic accuracy.

Beyond full-sentence descriptions, some datasets introduce more structured forms of annotation. RSITMD~\cite{yuan2022exploring}, for example, provides keyword-level annotations for 4,743 RGB images with spatial resolutions ranging between 0.3 and 1 meter. The dataset includes more than 23,000 keywords describing object attributes such as color, size, and spatial relationships, enabling downstream tasks such as image retrieval and fine-grained semantic interpretation.

Other efforts focus on more specialized captioning tasks. LEVIR-CC~\cite{liu2022remote} is a change-captioning dataset composed of 10,077 image pairs representing before-and-after observations of the same location. Each pair is associated with five human-written captions describing meaningful changes in the scene, such as building construction or deforestation, while ignoring irrelevant variations such as illumination or seasonal differences. The imagery is sourced from Google Earth and primarily covers urban and suburban regions in Texas, USA.

More recently, large-scale datasets combining automated and human-assisted annotation have emerged. XLRS-Bench~\cite{wang2025xlrs} introduces ultra-high-resolution remote sensing images accompanied by captions initially generated with GPT-4o. The annotation process follows a semi-automatic pipeline in which captions are first produced using a combination of sub-image analysis and global scene context, and are then refined by human annotators. These refinements correct object counts, improve factual accuracy, identify anomalies, and enhance sentence structure. The resulting dataset contains 934 long-form captions describing 1,400 images and represents one of the first efforts to integrate ultra-high-resolution imagery with large-scale caption generation and human quality control.

While most existing captioning datasets focus on optical RGB imagery, only a limited number include Synthetic Aperture Radar (SAR) data. The SAR Ship Image Captioning dataset~\cite{zhao2022exploring} contains 1,500 SAR images acquired from Sentinel-1 and TerraSAR-X satellites, annotated with 3,000 human-written captions. The descriptions emphasize the presence of ships and their spatial context while explicitly avoiding vague size descriptors and numerical ship counts.

The SARLANG-1M dataset~\cite{wei2025sarlang} represents a large-scale benchmark composed of one million SAR images and 31,968 captions automatically generated by three vision-language models: BLIP, CLIP, and GPT-4o. The dataset uses paired SAR and RGB imagery as input to generate captions with varying levels of descriptive detail, aiming to capture global scene characteristics and multimodal information.

Another recent contribution is the SAR-TEXT dataset~\cite{cheng2025sar}, which includes more than 130,000 SAR image–caption pairs generated using the SAR-Narrator framework~\cite{cheng2025sar}. This framework adopts a multi-stage captioning strategy to produce textual descriptions of SAR imagery and is designed to bridge the gap between SAR data and vision–language learning. The dataset supports the training and evaluation of several multimodal models across different tasks. Although the dataset includes images with multiple resolutions, the medium-resolution subset focuses exclusively on maritime scenes containing ships.
Another recent large-scale contribution is BigEarthNet.txt~\cite{herzog2026bigearthnet}, a multi-sensor image-text dataset built on co-registered Sentinel-1 SAR and Sentinel-2 imagery. It contains 464,044 image pairs and approximately 9.6 million text annotations, including captions, VQA pairs, and referring expression detection tasks. Annotations are generated through a pipeline combining template-based descriptions derived from land cover maps with LLM-based augmentation, ensuring semantic grounding and linguistic diversity. A manually verified benchmark split is also provided. While BigEarthNet.txt significantly advances large-scale multimodal learning in remote sensing, its captions are not fully human-written, as they are produced through semi-automatic processes.

More recently, large-scale vision–language datasets have been introduced. Among them, the GAIA dataset~\cite{zavras2026gaia} provides 40,201 remote sensing images paired with five automatically generated captions each, obtained through a pipeline combining web-scraped data and GPT-4o. The dataset covers a wide range of sensors, spatial resolutions, and geographic regions, enabling applications such as captioning, retrieval, and classification.
Although GAIA is described as multimodal, this refers to the diversity of sensing modalities across the dataset rather than aligned multimodal pairs. Each sample typically corresponds to a single modality, and all captions are synthetic rather than human-written.

Despite these recent advances, a significant gap remains in the current landscape. While recent large-scale datasets introduce multimodality, scale, and diverse annotation types, they often rely on automatically generated or semi-automatically refined captions and are not specifically designed for consistent multimodal alignment with purely human-written descriptions. Existing SAR captioning datasets remain domain-specific and typically focus on maritime scenarios, while most optical datasets rely on high-resolution imagery and lack multimodal pairing. 

To the best of our knowledge, no publicly available dataset provides fully human-written captions for jointly aligned SAR and optical imagery at medium resolution. This limitation restricts the development of models capable of robust multimodal scene understanding under realistic Earth observation conditions. In this context, Sentinel2Cap is designed to fill this gap by combining co-registered SAR and optical data with human-annotated captions at medium spatial resolution, enabling more reliable and interpretable multimodal learning.

\section{Dataset}
\subsection{Image source}
We use data from reBEN~\cite{clasen2024reben}, a large-scale multimodal remote sensing dataset that includes 549,488 pairs of Sentinel-1 and Sentinel-2 image patches. Each patch measures $120\times 120$ pixels at a spatial resolution of 10 meters and is sourced from 10 different European countries. reBEN extends the BigEarthNet (BEN) dataset~\cite{sumbul2019bigearthnet} by incorporating pixel-wise segmentation masks for every image, derived from the 2018 CORINE Land Cover (CLC) map\footnote{https://land.copernicus.eu/content/corine-land-cover-nomenclature-guidelines/html/}. The CLC offers a hierarchical classification with three levels of detail; for segmentation tasks, reBEN adopts the most detailed Level 3 (L3) scheme, which distinguishes 44 land cover classes. However, for image-level labels, it maintains the 19-class system originally used in BEN. Furthermore, reBEN introduces a new data splitting strategy based on geography, designed to minimize spatial overlap between training, validation, and test regions.

From reBEN 12,000 images are selected to keep a good balance in the number of classes present in the image.

\subsection{Image pre-processing}
To generate optical RGB images from the reBEN dataset, Bands 4, 3, and 2 (corresponding to red, green, and blue) are stacked to form natural-color images. This 3-channel representation is adopted both to provide an intuitive visualization used during the human annotation process for caption generation, and to ensure compatibility with the vision-language model used in this work, which expects 3-channel inputs. The resulting images are linearly normalized before being used as model inputs.

To generate a 3-channel representation of multi-spectral images, Bands 6, 8A  and 12 are selected, stacked and normalized to create a pseudo-RGB image. These bands are chosen to capture complementary spectral information: Band 6 and Band 8A are sensitive to vegetation vigor and chlorophyll content, supporting plant health and biomass assessment, while Band 12 is sensitive to moisture content and is useful for soil and vegetation analysis. Since all three bands have a spatial resolution of 20 m, no resampling is required. The use of three channels also follows the same rationale as for RGB images, ensuring compatibility with the model input requirements and maintaining consistency with the representation used for annotation. Together, they provide a diverse spectral coverage across the near-infrared and shortwave-infrared regions.

For SAR images, a percentile-based clipping strategy is applied to reduce the influence of extreme backscatter values in dB. The 0.1\% and 99.9\% percentiles are computed independently for each polarization channel to preserve their distinct statistical distributions. The resulting ranges were [-29.02, 4.20] for VV and [-33.97, -4.28] for VH. After clipping, each channel is linearly normalized to the [0, 1] interval.
A third channel is generated from the previous polarization channels by computing their difference (VV - VH), since the values are expressed in dB. This operation corresponds to a logarithmic ratio and highlights structural scattering differences between the two polarizations. The same clipping and normalization procedure is applied (range [-3.94, 19.43]). Finally, a three-channel pseudo-RGB representation is constructed using VV as red, VH as green, and VV-VH as blue. This design choice is motivated by prior findings~\cite{tosato2024can} showing that representing SAR data with three channels instead of two can improve feature extraction in convolutional neural networks; we adopt the same rationale here while also ensuring compatibility with models expecting 3-channel inputs.

\subsection{Month Distribution}
In addition, certain regions were captured multiple times by the satellite, often under varying weather conditions such as cloud cover, haze, or snow, which can reduce the visibility of ground features. To address this, a filtering step was applied to retain only one image per area, prioritizing acquisitions from months with lower average precipitation to ensure better visual quality. As a result, no duplicate areas are present in the final dataset.

\begin{figure}[!t]
\centering
\includegraphics[width=\columnwidth]{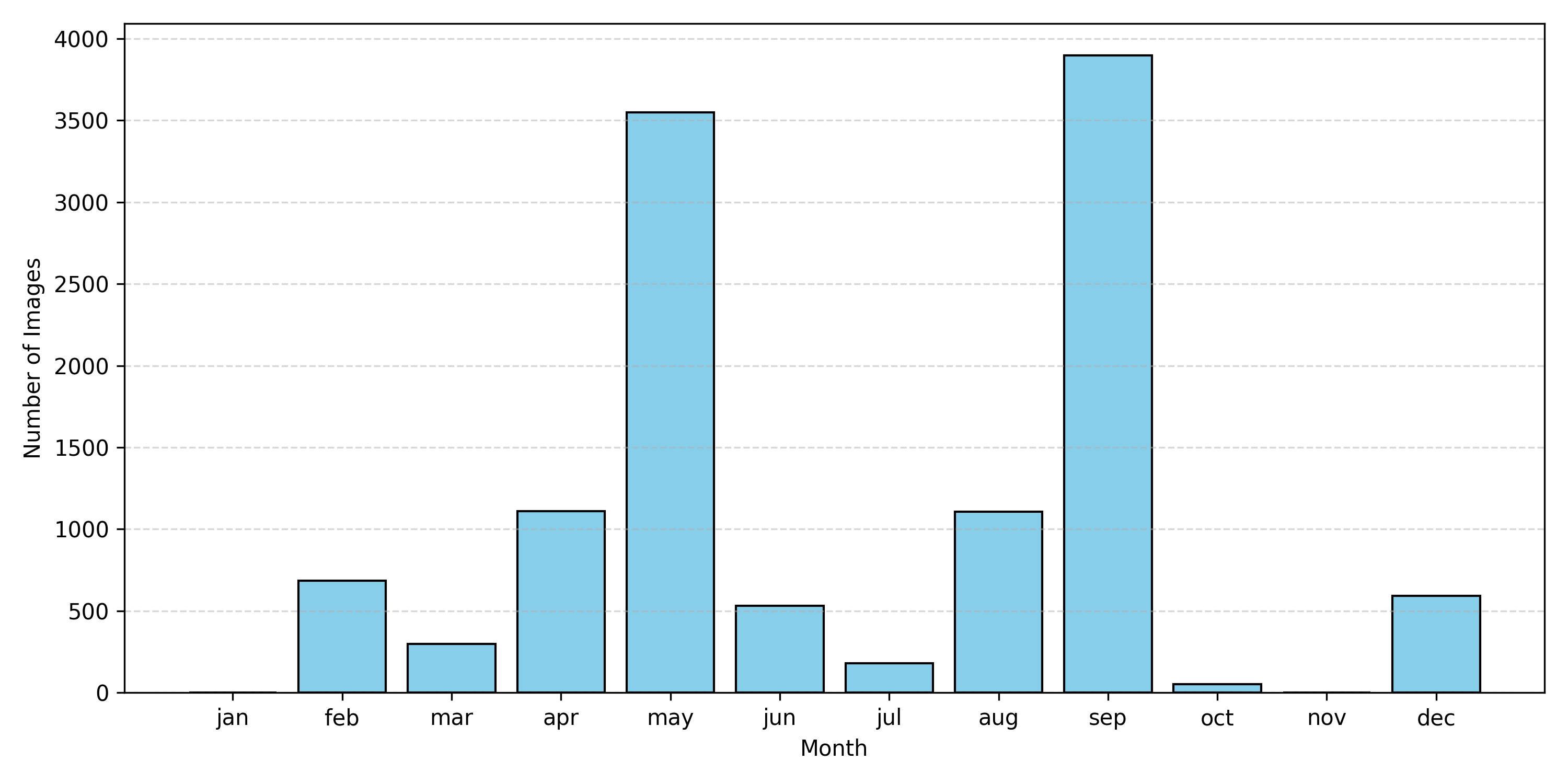}
\caption{Month Occurrence of \textit{Sentinel2Cap}}
\label{fig:monthly_distribution}
\end{figure}

The final monthly distribution of selected samples is shown in Figure~\ref{fig:monthly_distribution}. The choice of acquisition months reflects an effort to maximize image clarity while maintaining temporal diversity across the dataset.

\subsection{Distribution of the number of classes per image}
We retain the full Level 3 nomenclature with 44 land cover classes to maximize variability and increase the richness of the descriptions. However, rather than focusing on the frequency of individual semantic classes, we control the distribution of the \emph{number of classes per image}, which directly impacts caption complexity.

Images are selected such that the number of classes per image follows a Gaussian-like distribution. This means that most images contain a moderate number of classes, while images with very few or very many classes are less frequent. This choice provides a structured and realistic balance: it avoids an over-representation of overly simple scenes, while still limiting the number of highly complex scenes that may be difficult to describe consistently.

This selection strategy encourages the generation of captions with varying levels of complexity, promoting diversity in spatial relationships and semantic content. In turn, this helps captioning models learn to describe both simple and moderately complex scenes without being biased toward trivial or overly cluttered examples.

To ensure high-quality captions, images containing only a single class are deliberately excluded, as they tend to produce overly short and uninformative descriptions lacking spatial relationships and semantic richness. The underlying assumption is that a model capable of describing multi-class scenes with spatial interactions can generalize to simpler single-class scenarios.

\begin{figure}[!t]
\centering
\includegraphics[width=\columnwidth]{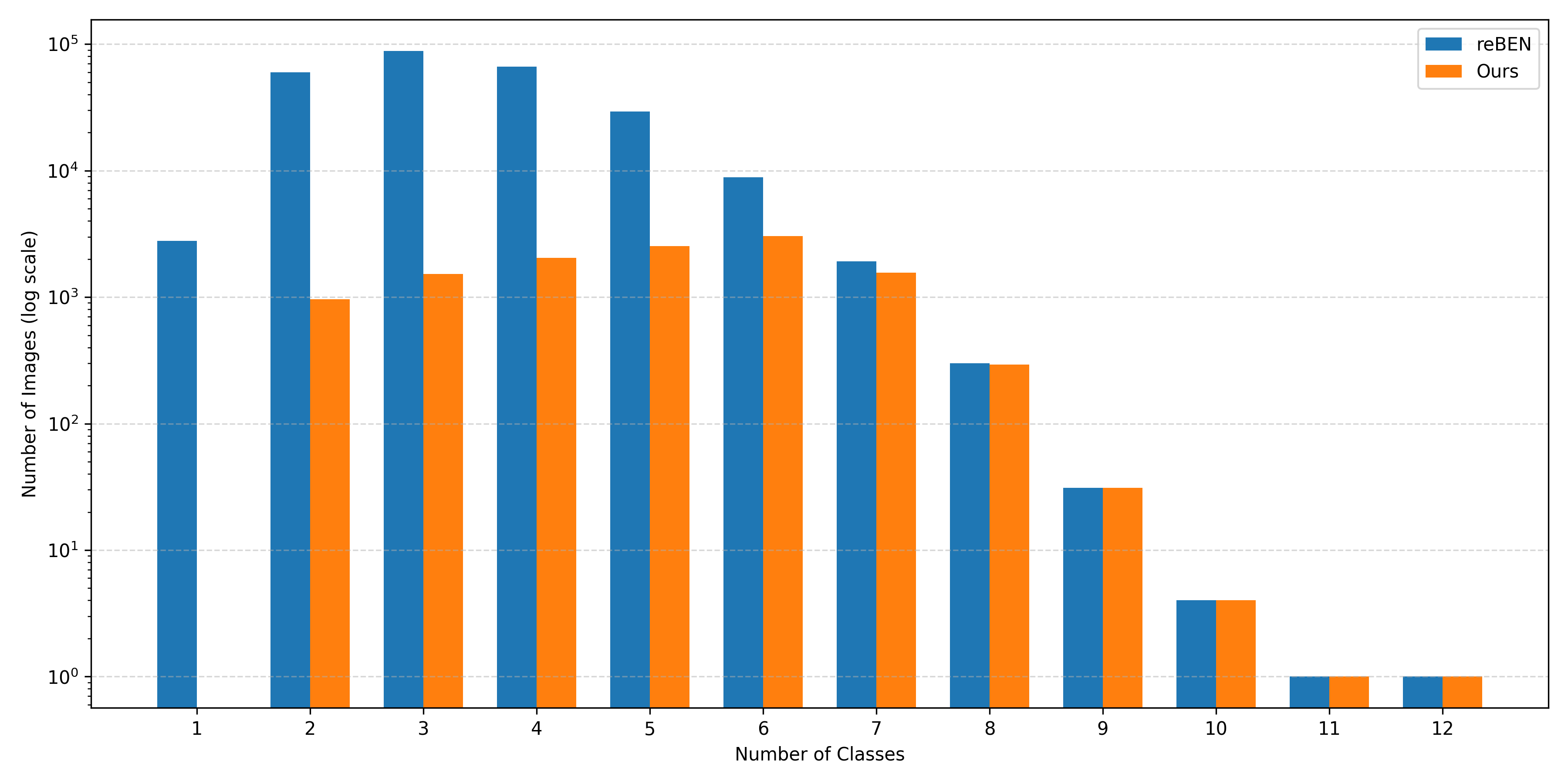}
\caption{Class distribution (log scale) comparison between reBEN and Sentinel2Cap.}
\label{fig:class_distribution_logscale}
\end{figure}

Figure~\ref{fig:class_distribution_logscale} illustrates the class distribution of \textit{Sentinel2Cap}, with reBEN shown for reference. The proposed selection results in a smoother, bell-shaped distribution of the number of classes per image, promoting diversity in caption structure while avoiding extreme cases.

Table~\ref{tab:normality_tests} reports three standard normality tests applied to the class distribution of \textit{Sentinel2Cap}. The D'Agostino-Pearson test evaluates skewness and kurtosis jointly, the Shapiro-Wilk test measures agreement with a normal distribution based on ordered samples, and the Anderson-Darling test emphasizes deviations in the tails. 

Across all tests, \textit{Sentinel2Cap} shows statistics and p-values indicating a distribution closer to normality, although normality is not strictly satisfied. This is consistent with the visual trend observed in Figure~\ref{fig:class_distribution_logscale}.

Overall, the results suggest that the proposed selection strategy produces a more balanced and well-structured class distribution, which is beneficial for generating diverse and consistent captions.

\begin{table}[ht]
\centering
\caption{Summary of Normality Test Results for Class Distributions}
\label{tab:normality_tests}
\begin{tabular}{lccc}
\toprule
\textbf{Test} & \textbf{Metric} & \textbf{reBEN} & \textbf{Ours} \\
\midrule
D'Agostino-Pearson & $S$ / p-value & 4.06 / 0.132 & \textbf{1.93} / \textbf{0.381} \\
Shapiro–Wilk       & $S$ / p-value & 0.722 / 0.0014 & \textbf{0.844} / \textbf{0.031} \\
Anderson–Darling   & $S$ / $C$ (5\%) & 1.535 / \textbf{0.679} & \textbf{0.761} / \textbf{0.679}\\
\bottomrule
\end{tabular}
\end{table}

\subsection{Class Distribution}
\begin{figure}[!t]
\centering
\includegraphics[width=\columnwidth]{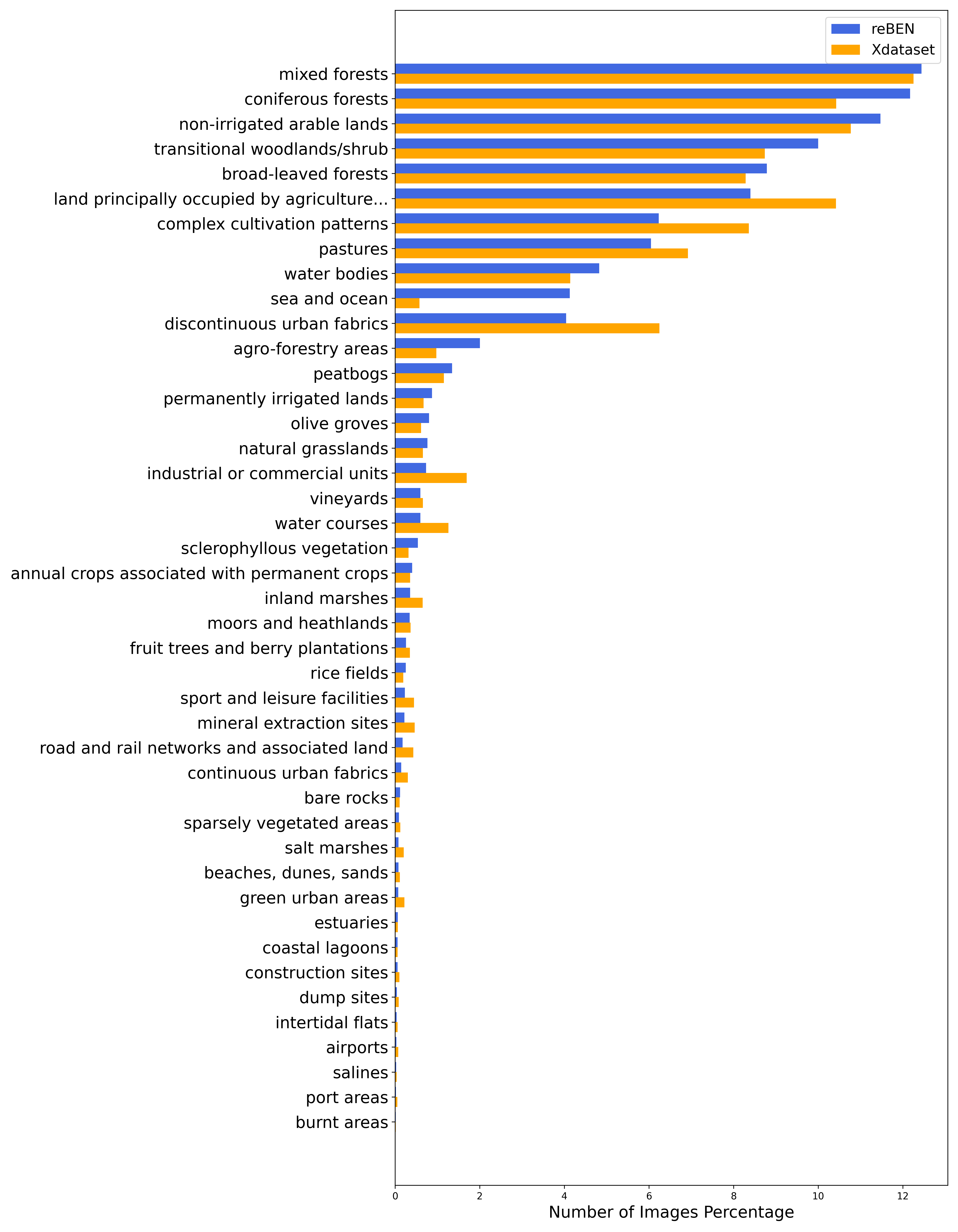}
\caption{Comparison of Class Occurrence Between reBEN and \textit{Sentinel2Cap}}
\label{fig:class_occurrence_comparison}
\end{figure}

The correlation coefficient between the class distributions of reBEN and \textit{Sentinel2Cap} is very high at 0.9696, indicating a strong positive linear relationship. This means the overall pattern of how classes are represented across the two datasets is very similar (see also Figure~\ref{fig:class_occurrence_comparison}). 

The mean absolute difference in percentage points, which measures the average absolute gap in class proportions between the two datasets, is effectively zero (0.0000). This indicates that, on average, the difference in the class proportions between the datasets is negligible when measured in absolute terms. However, when considering relative changes, the mean relative difference, which captures the average proportional change with respect to the baseline distribution in reBEN, rises to 37.27\%. This highlights that while the datasets generally align, some classes exhibit substantial proportional differences relative to their baseline frequencies in reBEN. The maximum absolute difference, defined as the largest absolute discrepancy observed for any single class, is 2.2049\%  points, observed for class ''discontinuous urban fabrics``, which reflects the largest absolute discrepancy but still remains small. On the other hand, the largest relative difference is much more pronounced at 191.88\%, occurring for class ''green urban areas``, indicating this class is nearly twice as prevalent in \textit{Sentinel2Cap} compared to reBEN.

Specifically, ''class green urban areas`` shows the largest over-representation in Sentinel2Cap, appearing at nearly triple its proportion in the original dataset. Conversely, class ''sea and ocean'' is the most under-represented, with its share reduced by about 86.01\%. This is mainly because around 83.45\% of the images labeled as ``sea and ocean'' in reBEN contain only that single class. Since \textit{Sentinel2Cap} excludes images with just one class, this category is substantially less prevalent, resulting in its significant under-representation compared to reBEN.

In summary, despite the very close overall distributions between the two datasets, some classes are distinctly over- or under-represented in Sentinel2Cap, reflecting its smaller size and possible selection biases. The strong correlation and minimal average absolute difference affirm the datasets’ general similarity, while the relative differences provide insight into specific class-level variations.

\begin{table}[ht]
\centering
\caption{Comparison statistics between class distributions in \texttt{reBEN} and \texttt{Sentinel2Cap}. Class 112 corresponds to ''discontinuous urban fabrics``, class 141 represents ''green urban areas``, and class 523 denotes ''sea and ocean``. }
\label{tab:class_dist_stats}
\begin{tabular}{ll}
\toprule
\textbf{Metric} & \textbf{Value} \\
\midrule
Correlation coefficient & 0.9696 \\
Mean absolute difference (percentage points) & 0.0000 \\
Mean relative difference (\%) & 37.27 \\
Maximum absolute difference (percentage points) & 2.2049 (class 112) \\
Maximum relative difference (\%) & 191.88 (class 141) \\
Largest over-representation in \texttt{Sentinel2Cap} & class 141 (191.88\%) \\
Largest under-representation in \texttt{Sentinel2Cap} & class 523 (-86.01\%) \\
\bottomrule
\end{tabular}
\end{table}

The balance of class distributions in reBEN and \textit{Sentinel2Cap} was evaluated using three metrics: the Coefficient of Variation (CV), Gini coefficient, and Shannon Entropy. The CV measures relative variability in class proportions, with lower values indicating more uniform distributions; here, \textit{Sentinel2Cap} has a slightly lower CV (1.58) than reBEN (1.60), suggesting a marginal improvement in uniformity. The Gini coefficient quantifies inequality where values near zero reflect more equality; again, \textit{Sentinel2Cap} scores lower (0.71 vs. 0.73), indicating a slightly more balanced class presence. Lastly, Shannon Entropy reflects diversity and evenness, with higher values denoting greater class representation uniformity; \textit{Sentinel2Cap} surpasses reBEN (2.77 vs. 2.75), supporting the same conclusion. Overall, these consistent results imply that \textit{Sentinel2Cap} exhibits a modestly more balanced distribution of classes, likely influenced by filtering choices such as the exclusion of single-class images, which affect the relative representation of certain classes.

\begin{table}[htbp]
\centering
\caption{Comparison of Class Distribution Balance Metrics Between \texttt{reBEN} and \texttt{Sentinel2Cap}}
\begin{tabular}{lcc}
\toprule
\textbf{Metric} & \textbf{reBEN} & \textbf{Sentinel2Cap} \\
\midrule
Coefficient of Variation (CV) & 1.60 & \textbf{1.58} \\
Gini Coefficient & 0.73 & \textbf{0.71} \\
Shannon Entropy & 2.75 & \textbf{2.77} \\
\bottomrule
\end{tabular}
\end{table}

\subsection{Caption Creation}
The captioning process is conducted by a dedicated team of approximately five to six individuals, each contributing to the manual annotation of satellite imagery. Each caption includes details about color, shape, and spatial relationships, incorporating both absolute references such as "in the top right of the image" and comparative relationships such as "to the right of the water body." The segmentation maps are not the primary source of information but are used as a reference, particularly to verify class names when there is uncertainty. When elements appear in the Sentinel-2 images that are not represented in the segmentation maps, annotators still include them to ensure completeness.

On average, the team spends twenty to twenty-five minutes per image, reflecting the careful attention given to each annotation. To ensure quality and consistency, the dataset undergoes a \textbf{Quality Assurance and Quality Control (QAQC) process}:  

\begin{itemize}
    \item \textbf{Quality Assurance (QA)}: The annotation workflow is designed to promote accuracy from the start. Annotators follow strict guidelines to describe all relevant visual elements, verify class names against segmentation maps when necessary, and maintain consistent use of spatial and descriptive terms.
    \item \textbf{Quality Control (QC)}: Each caption undergoes multiple verification steps. First, a manual review checks that all referenced classes match the visible content and resolves any uncertainties using segmentation maps. Second, a grammar review corrects language-related issues. Finally, an automatic check validates spelling, spacing, and capitalization of words without altering the descriptions or their meaning, ensuring semantic integrity is preserved.
\end{itemize}

This multi-level QAQC procedure ensures that the dataset is both semantically accurate and linguistically polished, supporting high-quality captioning for downstream tasks.

\subsection{Caption Length and Lexical Diversity}
\begin{figure*}[!t]
\centering
\includegraphics[width=\textwidth]{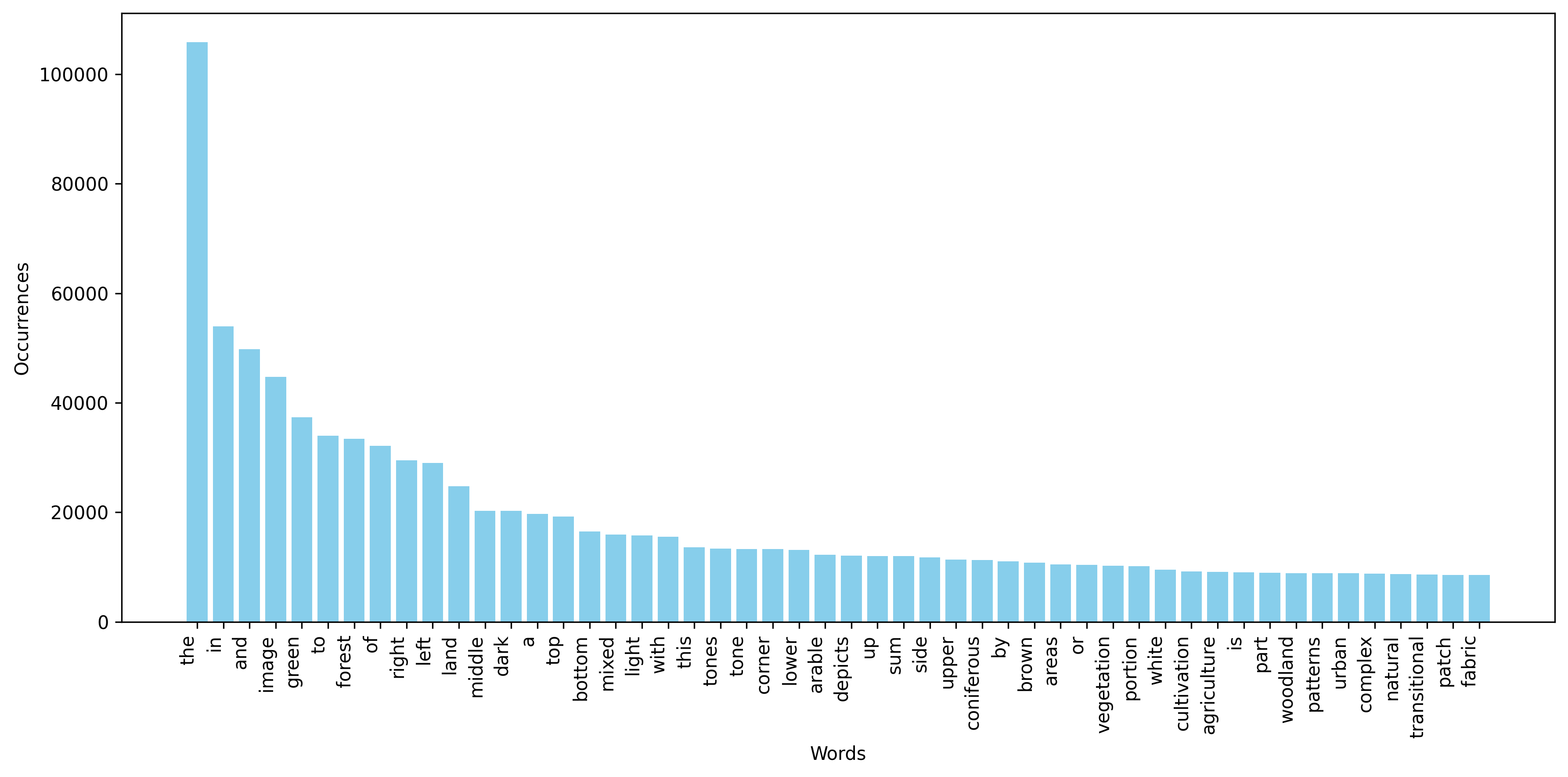}
\caption{50 most present words in \textit{Sentinel2Cap} dataset.}
\label{fig:50mostpresentwords}
\end{figure*}

The dataset contains 1'323'580 words, 1'837 unique words, each image has an average use of 720.51 occurrences, and a median of 6. The average size of the sentences not considering punctuation is 112.30 characters, while the median is 114.0. 
Figure~\ref{fig:50mostpresentwords} shows the 50 most present words in the dataset. 

\section{Model}
Qwen3-VL-8B-Instruct~\cite{qwen3technicalreport} is a large-scale multimodal vision-language model with approximately 8 billion parameters. It is designed to jointly process visual and textual inputs, enabling tasks such as image captioning, visual question answering, document understanding, and fine-grained scene description. The model is selected as a strong general-purpose baseline due to its competitive performance, moderate computational requirements, and public availability, which ensures reproducibility. Moreover, using a general-domain vision-language model allows us to better assess the impact of modality-specific prompting without relying on architectures explicitly tailored to remote sensing data. The model integrates a vision encoder with a large language model through a cross-modal alignment mechanism, allowing it to generate coherent and contextually grounded textual outputs conditioned on image content.

The “Instruct” version is specifically fine-tuned using instruction-following data, which improves its ability to respond accurately to natural language prompts. This makes it particularly suitable for zero-shot or prompt-based evaluation settings, where no task-specific fine-tuning is performed.

In our baseline setting, the model was used in a zero-shot manner to generate textual descriptions from the input images in \textit{Sentinel2Cap} .

Two different prompts were used. The first, referred to as the \textbf{base prompt}, states:

\textit{Describe this satellite image in one single continuous paragraph comprising less than 200 words. Do not use bullet points, numbered lists, or section titles. Provide a detailed and natural description focusing on land cover, structures, spatial layout, and colors. Be precise about the location of the different elements in the image, for example in the upper portion or in the middle. Use the following format: In this image, [description]. To sum up, [summary].}

The second is the \textbf{modality-specific prompt} and provides additional information about the satellite source and the image channels to give more context to the model. It states:

\textit{Describe this satellite image in one single continuous paragraph comprising less than 200 words. Do not use bullet points, numbered lists, or section titles. The image is taken from the  satellite \{\} and uses the following channels: \{\}. Use this information to better understand the image, but do not include it in the description. Provide a detailed description focusing on land cover, structures, spatial layout, and colors. Be precise about the location of the different elements in the image, for example in the upper portion or in the middle. Use the following format: In this image, [description]. To sum up, [summary].}

The placeholders \{\} are filled according to the modality: with 'Sentinel-2' and 'B4 Red, 'B3 Green, B2 Blue' for Sentinel-2 RGB images; with 'Sentinel-1' and 'VV Red, VH Green, VV/VH ratio Blue' for Sentinel-1 images; and with 'Sentinel-2' and 'B6 Red, 8A Green, B12 Blue' for Sentinel-2 images using 20 m resolution bands.

\section{Evaluation Scores}
To evaluate the quality of the generated captions, we used five widely recognized metrics: BLEU, METEOR, ROUGE and CIDEr. Each metric captures different aspects of similarity between the predicted and reference captions.

\subsection{BLEU} (BiLingual Evaluation Understudy) is a precision-based metric that calculates the overlap of $n$-grams between the candidate and reference captions. The BLEU score is computed as:
\begin{equation}
\text{BLEU} = \text{BP} \cdot \exp\left( \sum_{n=1}^{N} w_n \log p_n \right)
\end{equation}
where $p_n$ is the modified $n$-gram precision, $w_n$ is a weight (typically uniform), and BP is the brevity penalty to penalize overly short outputs. BLEU is widely used for machine translation but does not account for synonyms or recall.

In our case, we report BLEU-1, BLEU-2, BLEU-3, and BLEU-4, which correspond to using $N = 1, 2, 3,$ and $4$ respectively, thus progressively evaluating unigram, bigram, trigram, and four-gram overlap between the generated and reference captions.

\subsection{METEOR} (Metric for Evaluation of Translation with Explicit ORdering) improves upon BLEU by considering both precision and recall, aligning words using exact matches, stems, synonyms, and paraphrases. It also penalizes disordered word sequences. The METEOR score is calculated as:
\begin{equation}
F_{\text{mean}} = \frac{10 \cdot P \cdot R}{R + 9P}
\end{equation}
with additional penalties based on fragmentation of matched words. This metric often correlates better with human judgments than BLEU.

\subsection{ROUGE} (Recall-Oriented Understudy for Gisting Evaluation) is typically used for summarization tasks and focuses on recall how many reference $n$-grams are present in the candidate. ROUGE comes in several variants, including ROUGE-N (based on $n$-gram recall) and ROUGE-L (based on the longest common subsequence). For our evaluation, we relied on the ROUGE-L definition.

\subsection{CIDEr} (Consensus-based Image Description Evaluation) is designed for image captioning. It compares the candidate caption against multiple references, weighting $n$-grams by TF-IDF to emphasize informative terms. The score is computed as:
\begin{equation}
\text{CIDEr}(c, S) = \frac{1}{N} \sum_{n=1}^{N} \text{CIDEr}_n(c, S)
\end{equation}
where $c$ is the candidate caption, $S$ is the set of reference captions, $n$ denotes the $n$-gram order (e.g., unigrams, bigrams, etc.), and $N$ is the maximum $n$-gram length considered (typically $N=4$). CIDEr captures consensus across references and gives higher scores to rare but relevant words.

\section{Results and Discussion}

\begin{figure*}[!t]
\centering
\includegraphics[width=\textwidth]{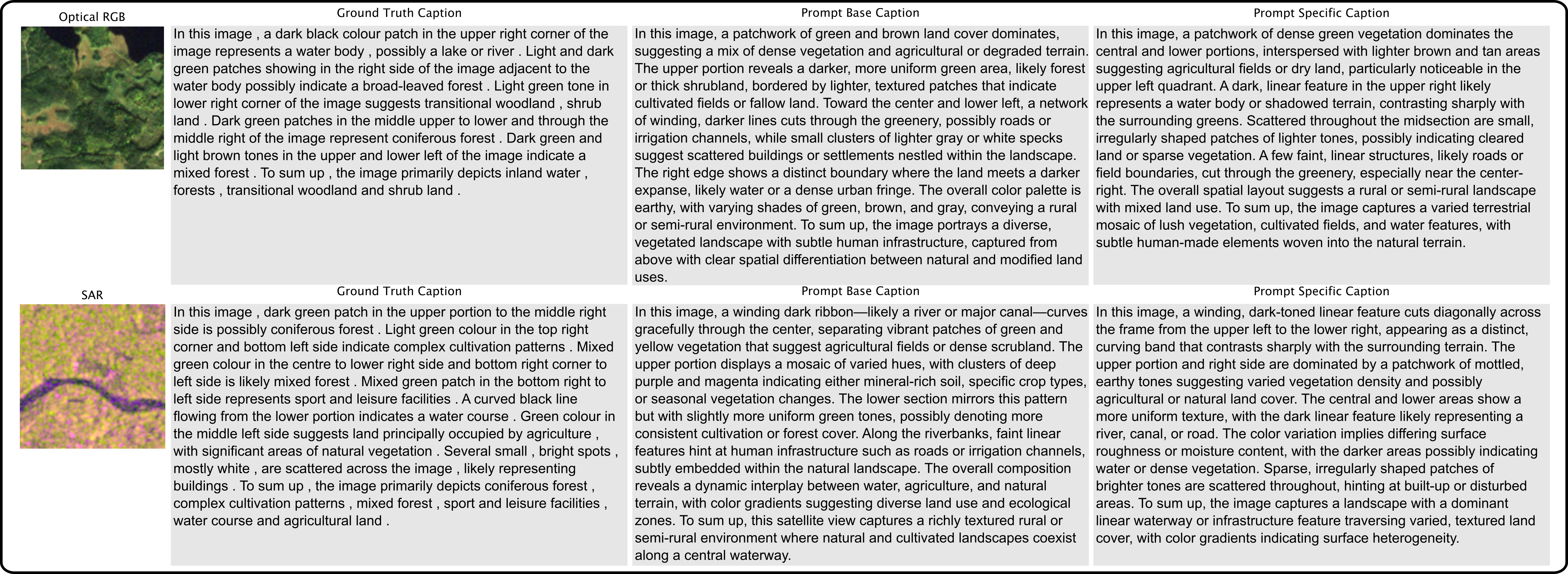}
\caption{Examples of the Qwen3-VL-8B-Instruct results on a RGB optical and SAR image.}
\label{fig:results}
\end{figure*}

\begin{table}[h]
\centering
\begin{tabular}{lccc}
\hline
& OPT-RGB & SAR & OPT-MUL \\
\hline
\multicolumn{4}{l}{\textbf{Base prompt}} \\
\hdashline
BLEU-1 & 0.2140 & 0.1939 & 0.2104 \\
BLEU-2 & 0.0905 & 0.0761 & 0.0856 \\
BLEU-3 & 0.0163 & 0.0082 & 0.0110 \\
BLEU-4 & 0.0018 & 0.0007 & 0.0011 \\
METEOR & 0.1659 & 0.1486 & 0.1621 \\
ROUGE\_L & 0.1555 & 0.1446 & 0.1550 \\
CIDEr & 0.0099 & 0.0077 & 0.0091 \\
\hline
\multicolumn{4}{l}{\textbf{Modality-specific prompt}} \\
\hdashline
BLEU-1 & \textbf{0.2178} & 0.1999 & 0.2112 \\
BLEU-2 & \textbf{0.0930} & 0.0831 & 0.0886 \\
BLEU-3 & \textbf{0.0168} & 0.0103 & 0.0133 \\
BLEU-4 & \textbf{0.0020} & 0.0011 & 0.0014 \\
METEOR & \textbf{0.1717} & 0.1544 & 0.1687 \\
ROUGE\_L & \textbf{0.1592} & 0.1508 & 0.1571 \\
CIDEr & \textbf{0.0096} & 0.0076 & 0.0084 \\
\hline
\\
\end{tabular}
\caption{Evaluation results with metrics as rows and modalities as columns}
\label{tab:results_inverted}
\end{table}

The results presented in Table~\ref{tab:results_inverted} indicate clear trends when comparing the performance of Qwen3-VL-8B-Instruct across different image modalities and prompt types. In general, OPT RGB images achieve the highest scores across almost all metrics. This is expected, as these are optical RGB images, which closely resemble the natural images that Qwen3-VL-8B has been primarily trained on. The model can therefore recognize colors, textures, and structures more easily, producing more accurate and fluent captions.

For OPT multi-spectral (OPT MUL) images, which use 20 m resolution bands, scores are slightly lower than RGB, reflecting the increased complexity of interpreting spectral bands that are less directly aligned with the model’s pretraining. Despite this, the modality-specific prompt improves the scores across BLEU-1 to BLEU-4, METEOR, ROUGE-L, and CIDEr, suggesting that providing the satellite and channel information allows the model to better contextualize the spectral data and generate more precise descriptions.

SAR images consistently have the lowest scores, especially for higher-order BLEU metrics. This is unsurprising because SAR imagery encodes physical backscatter information rather than optical color or texture, which Qwen3-VL-8B is not inherently familiar with. Here, the model struggles to generate sequences that closely match the reference captions. Nevertheless, even for SAR, the modality-specific prompt yields modest improvements. Moreover, the performance gap between SAR and multispectral imagery is significantly larger under the base prompt than under the modality-specific prompt, suggesting that providing explicit context about channels and modality helps the model reason more effectively about the physical information present and better reflect it in the generated textual description.

These quantitative trends are also reflected qualitatively in Figure~\ref{fig:results}. Modality-specific prompting produces captions that are slightly more concise, less narrative, and better aligned with remote sensing terminology, particularly for SAR imagery where descriptions rely more on physically grounded attributes such as texture, roughness, and linear structures. This qualitative improvement is consistent with the observed average score increases of +0.0023 for OPT-RGB, +0.0039 for SAR, and +0.0021 for OPT-MUL. SAR shows larger gains due to its lower baseline and the stronger impact of modality-specific priors that introduce essential physical cues. In contrast, RGB already aligns well with pretraining, while multispectral data remains only partially interpretable despite added context. 

However, despite these gains, the generated captions remain more generic and less structured than human annotations, often lacking a clear prioritization of the most salient features.

Across all modalities, BLEU-1 remains relatively high, showing that individual words are often correct, but BLEU-3 and BLEU-4 drop sharply, reflecting difficulty in generating exact n-gram sequences. METEOR and ROUGE-L remain more stable and benefit more noticeably from modality-specific prompts, highlighting the semantic gains from giving context. CIDEr scores, while low overall, also improve slightly with modality-specific information, showing the model can better capture informative or rare terms when guided.

In summary, providing modality-specific context consistently improves performance across all metrics, with the largest gains observed for non-RGB modalities, confirming that guiding the model to understand the type of data it is observing allows it to produce more accurate and semantically meaningful captions. The results also underscore the impact of image modality: RGB optical images are the easiest for the model, multi-spectral images are moderately challenging, and SAR images are the most difficult.

\section{Conclusion}
This paper presented \textit{Sentinel2Cap}, a human-annotated multimodal captioning dataset derived from the reBEN remote sensing dataset. The dataset contains Sentinel-1 SAR and Sentinel-2 optical image patches with detailed textual descriptions designed to capture land cover types, spatial relationships, and visual characteristics within the scenes. By excluding single-class images and promoting a balanced class distribution, the dataset encourages richer and more informative captions.

A baseline evaluation was conducted using the Qwen3-VL-8B-Instruct model in a zero-shot captioning setting across three modalities: optical RGB, optical multi-spectral, and SAR pseudo-RGB images. The model was selected as a strong general-purpose baseline due to its competitive performance, moderate computational requirements, and public availability, ensuring reproducibility. Moreover, using a general-domain vision–language model allows for a clearer assessment of the impact of modality-specific prompting without relying on architectures explicitly tailored to remote sensing data. However, this choice may disadvantage modalities such as SAR, for which domain-specific models could provide more suitable feature representations. The results demonstrate that optical RGB imagery yields the highest captioning performance, while SAR imagery remains significantly more challenging for vision-language models. Providing modality-specific prompts improves performance across all metrics, suggesting that contextual information about sensor type and channels helps the model better interpret the data.

Overall, the findings highlight the importance of human-written captions and multimodal datasets for advancing research in remote sensing image captioning. The proposed dataset provides a valuable benchmark for studying cross-modal scene understanding and for evaluating future vision–language models on SAR and optical satellite imagery. In particular, it offers a collection of captions with a high level of semantic and grammatical consistency, while maintaining a well-balanced distribution of classes and a relatively uniform caption length across the dataset. These characteristics make it especially suitable for training and evaluating models that require reliable, high-quality textual supervision.

Future work could focus on several directions. First, vision–language models could be fine-tuned directly on the dataset to better adapt them to the characteristics of remote sensing imagery. The dataset could also be expanded by including images at multiple spatial resolutions, potentially incorporating additional imagery sources.

Another promising direction is the use of the generated captions for Remote Sensing Visual Question Answering (RSVQA) tasks. For example, in ~\cite{tosato2025sar}, class names extracted from images are provided as contextual information to a large language model (LLM) to answer questions about remote sensing images, leading to improved performance. A similar strategy could be explored by providing the captions generated in this work as contextual input to the LLM. This could be evaluated using the dataset introduced in ~\cite{tosato2025checkmate}, which presents question–answer pairs for the reBEN dataset and is designed to exhibit very low bias.

Finally, future research could investigate multimodal fusion strategies that jointly leverage SAR and optical imagery. Integrating complementary information from both modalities may improve caption generation quality and lead to richer and more informative descriptions of remote sensing scenes.

\section{Acknowledgement}
We would like to thank the IEEE GRSS that founded this project. 

\bibliographystyle{IEEEtran}
\bibliography{bib}

\vfill

\end{document}